# DRSI-Net: Dual-Residual Spatial Interaction Network for Multi-Person Pose Estimation

( Shang Wu[a], Bin Wang[a,*] )

[a] *School of Communication and Information Engineering, Shanghai University, Shanghai, China*

[*] *Corresponding author: Bin Wang*

E-mail address: brantleywang@shu.edu.cn

# Abstract：

Multi-person pose estimation (MPPE), which aims to locate the key points for all persons in the frames, is an active research branch of computer vision. Variable human poses and complex scenes make MPPE dependent on local details and global structures; their absence may cause key point feature misalignment. In this case, high-order spatial interactions that can effectively link the local and global information of features are particularly important. However, most methods do not include spatial interactions. A few methods have low-order spatial interactions, but achieving a good balance between accuracy and complexity is challenging. To address the above problems, a dual-residual spatial interaction network (DRSI-Net) for MPPE with high accuracy and low complexity is proposed herein. Compared to other methods, DRSI-Net recursively performs residual spatial information interactions on the neighbouring features so that more useful spatial information can be retained and more similarities can be obtained between shallow and deep extracted features. The channel and spatial dual attention mechanism introduced in the multi-scale feature fusion also helps the network to adaptively focus on features relevant to the target key points and further refine the generated poses. Simultaneously, by optimising the interactive channel dimensions and dividing the gradient flow, the spatial interaction module is designed to be lightweight, thus reducing the complexity of the network. According to the experimental results on the COCO dataset, the proposed DRSI-Net outperforms other state-of-the-art methods in accuracy and complexity.

# 1 Introduction

A 2D multi-person pose estimation (MPPE), which predicts the key points of each person appearing in a given image, is an important computer vision task. MPPE has been widely used in pedestrian tracking [1], action recognition [2, 3], human-computer interaction [4] and image registration [5-7] because human poses can provide rich action and body structure information. In recent years, MPPE has made considerable progress with the powerful feature extraction capabilities of convolutional neural networks (CNNs) and transformers. However, it still has many challenges, such as multiple scales, occlusions of body parts, unknown numbers and changeable poses of persons. These uncertain factors seriously affect the accuracy of pose estimation.

## 1.1 Classification According to the Algorithmic Process

MPPE can be divided mainly into three categories according to the algorithmic process: top–down, bottom–up and recently proposed anchor-based methods. (1) Top–down methods [8-15] follow a two-stage strategy of 'detection first and estimation later'. They first use an object detector to detect each human instance from the input image and predict the key point positions for each person. This design leads to a linear increase in the computational complexity and memory consumption of the top–down method as the number of persons in the image increases. Furthermore, the top–down methods treat each person separately, ignoring the information clues of interactions between individuals. (2) In contrast, bottom–up methods [16-22] have less complexity and a constant inference time. Bottom–up methods first detect the positions of all human key points and group them into corresponding poses using various post-processing approaches. Bottom–up methods, such as OpenPose [16], PiPaf [17] and PPN [20], have focused on post-processing methods, which generally include pixel-level NMS, line integration, refinement, grouping and other steps. Although bottom–up methods have lower complexity, they typically have less accuracy because of their heavy reliance on the accuracy of post-processing grouping methods. In addition, bottom–up methods cannot be trained end-to-end because the post-processing steps occur outside the convolutional network and are not differentiable. Recently, some pioneers have solved these challenges and achieved good performance by decoupling poses [22] on a person. (3) Anchor-based MPPE, which directly regresses the key points relative to the anchor centre and achieves reasonable performance, was recently proposed by YOLOPose [23]. This method does not require the complicated post-processing procedures of bottom–up methods. Unlike top–down methods, in anchor-based methods, all people are localised together with their poses in a one-way inference process. Therefore, the complexity of anchor-based methods is independent of the number of people in the image. Compared to top–down and bottom–up methods, anchor-based methods demonstrate a more balanced performance in terms of accuracy and complexity.

## 1.2 Classification According to Technical Methods

If classified according to technical methods, MPPE can be divided into two categories: CNN-based methods and transformer-based methods. Although existing CNN-based methods [8-13, 16-22] have performed well in MPPE, their pose effects in complex scenes are poor because of overlapping, occlusion and various appearances and poses. Ever-changing human body poses and complex scenes make MPPE dependent on local details and global structure. Moreover, the absence of any of them may cause misalignments of the key point features. In this case, it is essential to have high-order spatial interactions [24] that can effectively connect local and global features. While CNN-based models have achieved significant results in human pose estimation, they lack long-term and high-order spatial interaction to capture the long-range dependencies of input features. To address this issue recently, TransPose [14] proposed using transformers [25] for human pose estimation. Using self-attention [25] with input-adaptive, long-range and high-order spatial interaction effectively captures long-range relationships in input features to improve pose-estimation performance. Inspired by HRNet [12], Yuan et al. proposed HRFormer [15] based on a multi-resolution parallel strategy and multi-head self-attention to improve accuracy. Although TransPose and HRFormer, two transformer-based models that rely on self-attention to capture long-term dependencies of features, have significantly improved pose-estimation accuracy, the high secondary complexity of self-attention also results in high complexity and memory consumption. Notably, the current transformer-based methods are all top–down methods. Top–down methods belong to the two-stage detection and estimation methods, and their inherent nature is relatively complex.

At present, there is a contradictory problem that existing CNN-based methods [8-12, 15-23] need high-order spatial interaction capabilities. Although transformer-based methods [14, 15] rely on the second-order spatial interaction capabilities of self-attention to optimise crowd and occlusion problems in complex scenes to a certain extent, they are not favoured by many real-time pose-estimation applications. This is because the high quadratic complexity of self-attention brings high computational complexity and memory consumption to transformer-based human pose estimation methods. Transformer-based methods and most existing CNN-based methods are also difficult to apply in real-time pose estimation because of their high complexity and slow inference speed.

## 1.3 Proposed Method

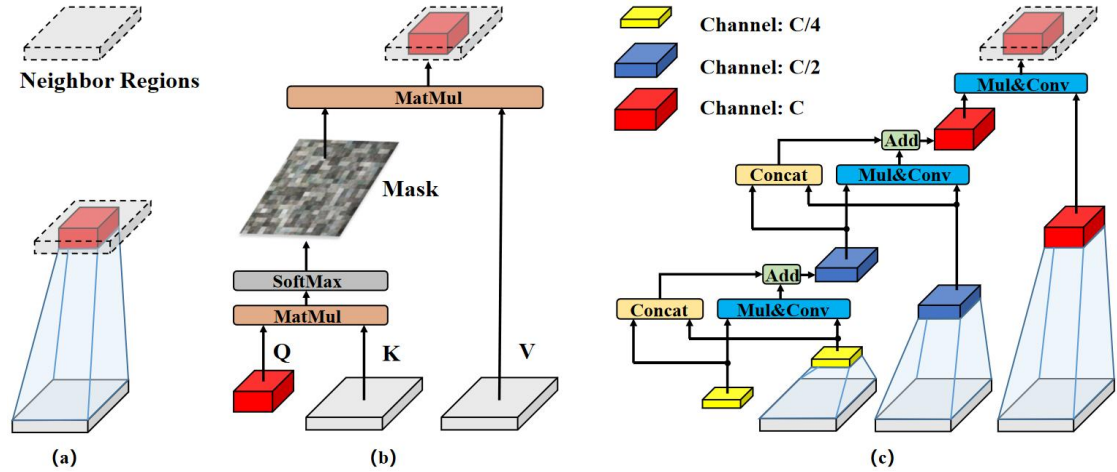

Figure 1: Comparison with existing pose-estimation methods. (a) Existing CNN-based methods utilise standard convolutions to extract features without spatial interaction. (b) The transformer-based methods perform two-order spatial interactions through self-attention. (c) The proposed method achieves high-order spatial interactions of a feature (yellow, blue and red) with the features of its neighbouring regions (grey) through recursive residual gated convolutions.

Based on the above problems, this paper proposes a novel anchor-based dual-residual spatial interaction network (DRSI-Net). This study enables CNN-based human pose estimators with high-order spatial interactions to achieve a good balance between accuracy and complexity.

To enable CNN-based MPPE with high-order spatial interactions, this paper proposed a recursive residual gated convolution (Res-g$^n$Conv) that focuses on the spatial interactions of features with neighbouring regions. With the large receptive field of a 7 × 7 depthwise convolution and the recursive residual spatial information interactions between features, Res-g$^n$Conv captures more global and local context information helpful for key point detection. Based on Res-g$^n$Conv, this paper proposes a DRSI block and a C3DR module with high-order spatial interaction and better representation ability. The continuously stacked DRSI blocks and C3DR modules continuously exchange information between features in a coarse-to-fine manner and learn the dependencies between features, providing high-quality pose information at different scales for the subsequent multi-scale feature fusion network (attentional spatial interaction path aggregation network, ASI-PAN). With the dual attention mechanisms of ASI-PAN, the proposed method can focus on meaningful ground-truth regions and key point positions rich in context information in features after high-order spatial interactions. In this way, multi-scale feature fusion can be performed efficiently, and the pose can be refined further at a finer level. Combining recursive residual spatial interactions and dual attention mechanisms optimises the key point misalignment issue and considerably improves performance.

Furthermore, on DRSI-Net, the channel of spatial interaction features is gradually increased, and the gradient flow of spatial interaction modules is divided. The above designs, combined with the anchor-based and end-to-end implementation, reduce the complexity of the proposed method compared to the state-of-the-art bottom–up and top–down methods.

Overall, the contributions of DRSI-Net are as follows:

- Res-g$^n$Conv, designed for the first time, solves the potential gradient explosion and performance degradation problems of ordinary recursive gated convolution [24]. It has similar capabilities to self-attention in transformers, e.g. robust long-range, input-adaptive and high-order spatial interactions, while avoiding the high quadratic complexity of self-attention.
- A DRSI block is further designed by combining a novel inverted bottleneck and a Res-g$^n$Conv. Compared to the transformer block [26], the DRSI block has higher-order spatial interactions and improves the representation ability of the network.
- To strengthen the learning ability of the DRSI block, reduce memory consumption and increase the network inference speed, a cross-stage DRSI (C3DR) module with segmented gradient flow and cross-stage feature fusion strategy is proposed.
- By stacking different numbers of C3DR modules and DRSI blocks and fusing Res-g$^n$Conv with the dual attention mechanism, this paper proposes a high-accuracy and low-complexity framework for MPPE. The proposed method recursively performs residual spatial interactions and continuously refines poses through attention.

The proposed method was evaluated using the challenging COCO dataset [27]. The method outperformed the other state-of-the-art methods. Specifically, DRSI-Net achieved better performance with 71.5% AP and 90.6% AP50 (on the COCO val2017 dataset) as well as 70.6% AP and 90.5% AP50 (on the COCO test-dev2017 dataset), with half the complexity of the SOTA bottom–up methods. These results highlight the effectiveness of DRSI-Net, and the aforementioned components are also efficient in ablation experiments.

# 2 Related Work

The existing 2D MPPE can be divided into three categories: top–down, bottom–up and recent anchor-based methods.

## 2.1 Top–down Methods

Top–down methods [8-15] are intuitive and effective by first detecting the bounding boxes of all people from a given image and then estimating the human poses. In 2017, GRMI [8] was proposed, which uses Faster RCNN [28] as a multi-person object detector to detect human bounding boxes. The model then used residual networks to extract feature information, predicting heatmaps and offsets for each human key point and further aggregating to obtain more accurate key point positions. Owing to the low accuracy of GRMI, RMPE [9] introduced the Symmetric Spatial Transform Network to address this issue. In addition, RMPE proposed the Parametric Pose NMS method to handle redundant results in the detection output. To address feature fusion and challenging key point detection, CPN [10] introduced GlobalNet based on the feature pyramid fusion module and RefineNet to correct the positions of difficult-to-detect key points. Simple Baseline [11] enhanced the key point detection accuracy by increasing the feature map resolution through transposed convolution. HRNet [12] used a multi-resolution parallel strategy to maintain the model at high resolution throughout the process, resulting in rich high-resolution features. The problem that CNN-based human pose estimators cannot capture long-range spatial dependencies was solved by TransPose [14], which proposed using a transformer [25] for human pose-estimation tasks and capturing the effective long-range relationship of input features through the self-attention in the transformer. HRFormer [15] combines the vision transformer [26] with multi-resolution parallel strategy proposed by HRNet, considerably improving the model performance. Although the top–down methods have higher accuracy, they first need to detect all persons through an object detection framework and then perform single-person pose estimation. When the number of persons increases, the accuracy of the object detection framework will decrease and the computational complexity will increase linearly.

## 2.2 Bottom–up Methods

The bottom–up methods [16-22] jointly detect the key points of everyone in the image, and the research focuses on how to accurately associate the detected key points with forming the corresponding human pose. OpenPose [16] enables the output of each stage to contain a 2D body part affinity field (PAF) and a confidence map. In the inference process, the connection between the key points can be judged only by the line integration of the PAF between the two candidate key points. Kreiss et al. proposed PifPaf [17], which is similar to OpenPose. PifPaf first locates each part of the human body through the part intensity field and then uses fine-grained PAF to associate these parts to form a complete human pose. Inspired by HRNet, Cheng et al. proposed HigherHRNet [21] through high-resolution parallel training and used associative embedding [18] to group the key points and improve detection accuracy. Owing to the negative impact of the high coupling of human key points in previous methods, DEKR [22] proposed the idea of decoupling key points. SOTA performance was achieved with the current bottom–up methods by performing feature extraction and regression on each key point independently. Compared to top–down methods, bottom–up methods are more suitable for real-time MPPE because of their fast inference speed and low cost, but their accuracy is slightly lower.

## 2.3 Anchor-based Method

The anchor-based YOLO frameworks [29-33] are fundamental in object detection. The recently proposed YOLOPose [23] is an end-to-end MPPE framework based on the popular YOLOv5 [33]. The framework provides a novel end-to-end, anchor-based method for 2D human pose estimation by directly regressing the key points relative to the anchor centre. The anchor point matching the ground-truth box stores the complete key point position, confidence and bounding box position of each person. Using heatmaps, distinguishing two similar key points from different human bodies that are spatially close together is challenging. However, if the two persons are matched by different anchors, it is easy to distinguish similar key points close to each other in space. The occlusion problem in MPPE has been optimised to some extent. Anchor-based methods avoid many non-differentiable post-processing operations commonly used in bottom–up methods. They group the detected key points into a skeleton because each detected box has an associated pose, making the key points inherently grouped. Unlike the high-cost top–down methods, in anchor-based methods, all people are localised with their poses in a one-way inference process. YOLOPose achieves competitive performance with SOTA bottom–up methods on the COCO dataset, demonstrating the significant research potential of anchor-based methods.

## 2.4 Spatial Interactions

The emergence of vision transformer (ViT) [26] has challenged the dominance of CNNs in computer vision. The critical factor for its success lies in a new approach to spatial modelling of two-order spatial interactions, long-range dependencies and input adaptations through self-attention. Direct global relationship modelling expands the receptive field of an image, thereby obtaining more contextual information. Several studies have been conducted based on standard convolutions to give CNN-based models these capabilities. Dynamic convolution [34, 35], the squeeze-and-excitation (SE) module [36] and the MSA block [37] endow convolution with the capability of additional spatial interactions by introducing dynamic weights. ConvNeXt [38] and RepLKNet [39] enlarge the receptive field by exploring larger convolution kernels. VAN [40] proposed large-kernel attention that generates different weights through large-kernel dilated convolutions and 3 × 3 standard convolutions and uses gated convolutions to obtain input-adaptive capability. PGFNet [5] uses a hierarchical class residual method and group space attention operation to process each feature group, improve the diversity of input features and capture more spatial information. Recently, HorNet [24] proposed a recursive gated convolution(gnConv) that uses an efficient combination of gated convolutions and recursive designs to implement the three key capabilities of ViT using a convolution-based framework. While extending the two-order spatial interactions in self-attention to any order, the convolution-based implementation also avoids the quadratic complexity of self-attention and the asymmetry brought about by local attention in transformer [25] and Swin transformer [41]. However, g$^n$Conv has potential gradient explosion and performance degradation problems as the interaction order and model depth increase. This paper proposes a Res-g$^n$Conv that solves the above problems and has more robust high-order spatial interactions. The proposed DRSI-Net, which combines the anchor-based method and the advantages of CNN and transformer, is a robust framework for MPPEs.

# 3 DRSI-Net

## 3.1 Overview of DRSI-Net

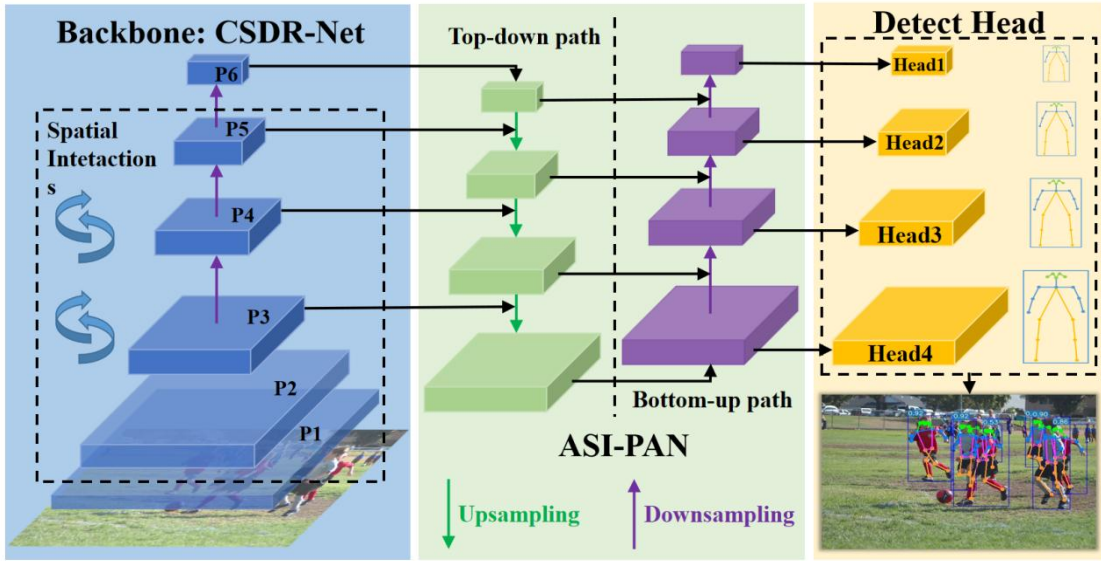

Figure 2: Architecture of a DRSI-Net. The input image first generates feature maps of different resolutions [P3, P4, P5 and P6] through the CSDR-Net backbone, and these four feature maps are then input into ASI-PAN for multi-scale feature fusion. Finally, the detection head is used to detect human bounding boxes and key points.

Figure 2 presents an overview of the DRSI-Net architecture, which comprises three parts from a macro perspective. First, as shown in the CSDR-Net backbone in Figure 5, the input is passed through a simple focus network to extract the shallow features of the RGB images. Four consecutively stacked downsampling convolutions (Conv) and cross-stage DRSI (C3DR) modules are then used to obtain feature maps of different scales [P2, P3, P4 and P5]. Conv comprises a 3 × 3 stride convolution with batch normalisation [42] and a sigmoid weighted linear unit (SiLU) [43]. This process is used to downsample the feature maps. The C3DR module comprises DRSI blocks based on Res-g$^n$Conv, which have robust arbitrary order, input-adaptive and long-range spatial interactions. The low-resolution feature map (P6) for small targets is obtained using the SPP and C3 modules [33]. The whole backbone is divided into five stages; the feature map resolution of each stage is halved compared with that of the previous stage. In the optimal implementation of DRSI-Net, the number of stacked DRSI blocks in the C3DR module of the four stages is [3, 9, 9 and 3]. Including 'Stage 5', the output channels of each stage are [128, 256, 512, 768 and 1024]. Through continuous interactive optimisation with the features of the neighbour regions, these features after continuous high-order spatial interactions will be rich in more useful contextual information.

After processing the backbone, the extracted P3, P4, P5 and P6 features are fed into ASI-PAN to fuse multi-scale features and refine the generated poses. ASI-PAN comprises two paths. The top–down path contains Res-g$^n$Conv with high-order spatial interactions, and the bottom–up path combines channel attention with spatial attention.

The ASI-PAN output is fed into detection heads corresponding to different object scales. Each detection head contains a bounding box head and a key point head that can simultaneously predict human bounding boxes and key point coordinates.

Res-g$^n$Conv is the most basic component of the entire framework, and the DRSI block, C3DR module and ASI-PAN are all based on it. Thus, the following content is divided into four parts to introduce them.

## 3.2 Recursive Residual Gated Convolution

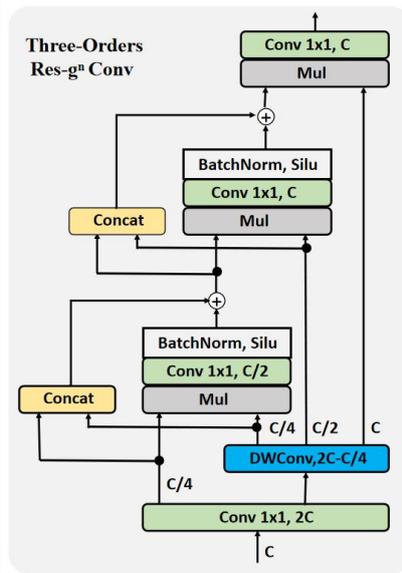

Figure 3: Illustration of recursive residual gated convolution (Res-g$^n$Conv), which has more robust high-order spatial interactions.

Section 2.4 briefly states that ViT has been considerably popular in computer vision tasks because it dynamically uses multi-head self-attention to generate weights for mixed spatial tokens. In contrast, CNNs only use static convolution kernels to capture the neighbouring features. Although multi-head self-attention inherits the powerful capabilities of self-attention, it also inevitably has high quadratic complexity. This severely limits the application of ViT, particularly in fields that require high resolution, such as object detection and human pose estimation. This paper proposes Res-g$^n$Conv with higher-order and lower-complexity spatial interactions than self-attention.

Gated convolution (gConv), which implements one-order input-adaptive interaction, is the basis of high-order interaction convolution. Let $x \in \mathbb{R}^{C \times H \times W}$ be the input feature of the gated convolution. The output can be written as follows:

$$[p_0^{C \times H \times W}, q_0^{C \times H \times W}] = \varphi_I(x) \in \mathbb{R}^{2C \times H \times W},$$
$$p_1 = p_0 \odot f(q_0) \in \mathbb{R}^{C \times H \times W}, \qquad (1)$$
$$y = gConv(x) = \varphi_O(p_1) \in \mathbb{R}^{C \times H \times W},$$

where $\varphi_I$ and $\varphi_O$ are the outputs of the fully connected linear projection layer after performing spatial mixing; $f$ is an ordinary depthwise convolution; $p_0$ and $q_0$ are obtained by splitting the input $x$ according to the channel dimension after a linear projection [24]. $p_0$ will perform a dot product operation with the neighbour feature $q_0$ after depth convolution processing to obtain $p_1$. $p_1$ will go through another linear projection layer to obtain the final output $y$. The above procedure shows that gated convolution explicitly introduces a one-order interaction between the neighbour features $p_0$ and $q_0$ through element-wise multiplication.

A significant gap exists between the one-order interaction of gConv and the two-order interactions achieved by two consecutive matrix multiplications in self-attention. To make CNNs have similar self-attention high-order spatial interactions, recursive gated convolution (g$^n$Conv) [24] is proposed. Let $x \in \mathbb{R}^{C \times H \times W}$ be the input feature of gnConv. The output of g$^n$Conv can be written as

$$[p_0^{C_0 \times H \times W}, q_0^{C_0 \times H \times W}, q_1^{C_1 \times H \times W}, \ldots, q_{n-1}^{C_{n-1} \times H \times W}] = f(x) \in \mathbb{R}^{(C_0 + C_q) \times H \times W},$$
$$p_k = p_{k-1} \odot f(q_{k-1}) \in \mathbb{R}^{C_1 \times H \times W}, 0 < k \leq n, \qquad (2)$$
$$y = g^n Conv(x) = \varphi_O(p_k) \in \mathbb{R}^{C \times H \times W},$$

where $\varphi_O$ is the output of the fully connected linear projection layer after performing spatial mixing; $f$ is a global filter or deep convolution; $n$ represents the number of recursions; $p_0$ and $\{q_k\}$ are the input after linear projection and channel dimension segmentation. First, $p_0$ performs a dot product operation with the adjacent feature $f(q_0)$ after processing by a global filter or deep convolution to obtain $p_1$. $p_1$ continues to perform a dot product operation with the adjacent feature $f(q_1)$ to obtain $p_2$. After $n$ iterations, the recursion ends and the final interaction feature $p_n$ is obtained, which is then passed through another linear projection layer to obtain the final output y. The above process introduces continuous interactions between adjacent features through element multiplication in g$^n$Conv.

g$^n$Conv successfully implements spatial interactions above second order through recursion and gated convolution, achieving good results. However, g$^n$Conv has potential gradient explosion and performance degradation problems when the interactive order and depth of the models are deepened. The proposed Res-g$^n$Conv solves the above problems through residual recursive designs and gives the model more robust high-order spatial interactions. Let $x \in \mathbb{R}^{C \times H \times W}$ be the input feature, similar to the input processing of g$^n$Conv. Res-g$^n$Conv first uses a linear projection to map its channel dimension to 2C. Through channel segmentation and a depthwise convolution $f$, a set of projected features $p_0$ and $\{q_k\}_{k=0}^{n-1}$ (referred to as a neighbour feature group of $p_0$) is obtained, as shown in Equation 3:

$$[p_0^{C_0 \times H \times W}, q_0^{C_0 \times H \times W}, q_1^{C_1 \times H \times W}, \ldots, q_{n-1}^{C_{n-1} \times H \times W}] = f(x) \in \mathbb{R}^{(C_0 + C_q) \times H \times W},$$
$$C_0 + C_q = 2C, \; C_q = \sum_{k=0}^{n-1} C_k, \; C_k = \frac{2C}{2^{n-k-1}}, \; 0 \leq k \leq n-1, \qquad (3)$$

where $C_q$ represents the total channel dimension of the neighbour feature group of $p_0$ and $C_k$ represents the channel dimension of each neighbour feature. $C_k$ controls the complexity of high-order interactions within an acceptable range by constraining the number of neighbour features divided by the depthwise convolution according to the channel dimension. This prevents excessive computational overhead caused by the high interaction order $n$. The lower-order interactions are assigned fewer channels, and the higher-order interactions are assigned more channels, i.e. the channel dimension of the interactive feature increases gradually and presents an inverted pyramid structure. Such a design realises coarse-to-fine information interactions and helps the network explore long-range dependencies.

Furthermore, depthwise convolution in Res-g$^n$Conv plays another vital role. Traditional CNNs usually use 3 × 3 convolutions in the entire network, while ViT computes multi-head self-attention within the entire feature map or a relatively large local window. A large receptive field can help the network capture more global semantic information. By setting the kernel size of depthwise convolution to 7 × 7, the gap of the receptive field between CNNs and ViT can be effectively decreased.

As shown in Figure 3, in contrast to the general identity connection, Res-g$^n$Conv concatenates the two neighbour features $p_k$ and $f_k(q_k)$ in the channel dimension to obtain the residual branch $\text{Res}_k$ (the last residual branch in Figure 3 is omitted):

$$\text{Res}_k = \text{Concat}(p_k, f_k(q_k)) \in \mathbb{R}^{C_{k+1} \times H \times W}, \tag{4}$$

where $\{f_k(q_k)\}$ is a set of neighbour features with gradually increasing dimensions obtained by depthwise convolution $f$ processing in Equation 3. During forward propagation, $p_k$ and $f_k(q_k)$ cooperate to form the residual branch, and the corresponding part is found for gradient calculation during backward propagation. The residual spatial interactions are then performed recursively on the obtained residual branch and the neighbouring feature group, as expressed in Equation 5. $\text{Res}_k$ and the interacted feature are connected by a skip connection to obtain the output of each order according to the following equation:

$$p_{k+1} = \left(\text{Res}_k + f_k(q_k) \odot g_k(p_k)\right) / \lambda, \quad k = 0, 1, \ldots, n-1, \tag{5}$$

where $\lambda$ is used to scale the output to stabilise training. After $p_k$ is processed by $g_k$ using Equation 6, its channel dimension is projected to the same dimension as $\{f_k(q_k)\}$ to realise the matching interaction with neighbour features at different orders:

$$g_k = \begin{cases} \text{Identity}, & k=0, \\ \text{SiLU}\left(\text{BN}\left(\text{Linear}(C_{k-1}, C_k)\right)\right), & 1 \leq k \leq n-1, \end{cases} \tag{6}$$

where BN is batch normalisation. After performing residual spatial interactions recursively, the obtained feature $p_n$ is input into the last linear projection layer to obtain the final result $\varphi_O$ of Res-g$^n$Conv. In the recursive interaction stage, the residual branch obtained by concatenating two neighbouring features, BN and SiLU, is introduced. This design solves the gradient explosion and performance degradation problems caused by relatively high-order spatial interactions and deep networks. It enhances the ability to perform high-order spatial interactions and improves the learning efficiency of the network. Although Res-g$^n$Conv realises high-order spatial interactions where $n$ is an arbitrary order, this does not mean that the order of interactions can be increased endlessly. Similar to the influencing factors, such as width and depth, the interaction order has different optimal value ranges for different models.

## 3.3 Dual-Residual Spatial Interaction Block

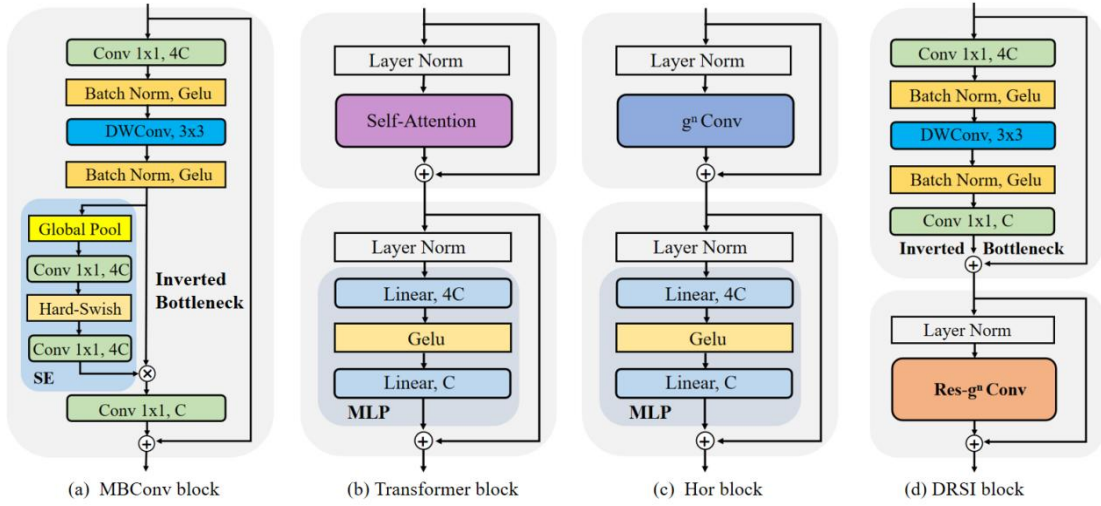

Figure 4: Comparison of the blocks. (a) The MBConv block consists of an inverted bottleneck and a squeeze-and-excitation (SE) module. (b) The transformer block consists of a self-attention module and an MLP module. (c) The Hor block consists of a $g^nConv$ and an MLP module. (d) The DRSI block proposed in this paper consists of a novel inverted bottleneck and a Res-$g^nConv$.

To improve the representation ability of the network, several blocks were proposed, such as the MBConv, transformer and Hor blocks. They were composed of different attention mechanisms and inverted bottlenecks. Mobile convolution (MBConv block) [44], which comprises an inverted bottleneck that is different from the bottleneck in ResNet [45] and an SE attention module [36], is shown in Figure 4(a). Figure 4(b) shows the transformer block [26], which contains a self-attention module and a multi-layer perceptron (MLP) module. The MLP module combined with the residual branch in the transformer block is a simple version of the inverted bottleneck. HorNet [24] successfully achieved a function similar to self-attention through $g^nConv$ and constructed a Hor block with the same meta-architecture as the transformer block based on $g^nConv$, as shown in Figure 4(c). The success of the above three blocks in computer vision tasks highlights the effectiveness of the combination of the attention mechanism and the inverted bottleneck in deep learning. To effectively utilise high-order spatial interactions of Res-$g^nConv$ and improve the representation ability of the network, this paper proposes a novel and efficient DRSI block.

The DRSI block contains a novel inverted bottleneck and Res-$g^nConv$ proposed in Section 3.2 (Figure 4(d)). Compared to the transformer block, the DRSI block replaces the MLP module with a new inverted bottleneck and replaces self-attention with Res-$g^nConv$, which has higher-order spatial interactions. Res-$g^nConv$ is a better alternative to self-attention and $g^nConv$ and can be called an attention mechanism. In particular, the order of the attention mechanism and the inverted bottleneck is reversed in the DRSI block. The new inverted bottleneck is calculated first and the final result is then obtained through Res-$g^nConv$.

In the inverted bottleneck of the DRSI block, a 1 × 1 standard convolution is used to increase the dimension of the input features, and a 3 × 3 depthwise convolution is used to extract the features. Finally, a 1 × 1 standard convolution reduces the dimensions to the original input dimensions. Assuming the input of the DRSI block is $x \in \mathbb{R}^{C \times H \times W}$, the output I(x) of the inverted bottleneck can be described as follows:

$$I(x) = x + (C_2 \circ D)(C_1(x)),$$
$$C_1(x) = \text{Conv}(\text{BN}(x)),$$
$$D(x) = \text{DW}\left(\text{GeLU}(\text{BN}(x))\right), \quad (7)$$
$$C_2(x) = \text{Conv}\left(\text{GeLU}(\text{BN}(x))\right),$$

where BN, DW and GeLU represent batch normalisation, depthwise convolution and Gaussian error linear unit [46], respectively. Replacing the MLP module with a novel inverted bottleneck adds more representation capabilities to the network. Inverted bottleneck blocks allow the network to learn residuals directly, which enables smoother feature transfer between layers and helps preserve and propagate useful information. In addition, inverted bottleneck blocks enable the network to learn residual mappings by introducing cross-layer connections, which improves the generalisation ability and training convergence speed of the model and helps alleviate the problems of vanishing and exploding gradients that occur during deep network training [44]. The inverted bottleneck delegates the downsampling task to its internal depthwise convolution, which helps the network learn better downsampling kernels and reduces the number of parameters and computational costs compared with standard convolutions.

After the output I(x) of the inverted bottleneck is obtained, the output of the entire DRSI block can be expressed as

$$y_{DR} = I(x) + \varphi_O\left(\text{LN}(I(x))\right), \quad (8)$$

where LN is layer normalisation [47] and $\varphi_O$ is the output of Res-g$^n$Conv. The DRSI block efficiently fuses the inverted bottleneck with Res-g$^n$Conv, which has a positive bottleneck. Based on the implementation of Res-g$^n$Conv, it can also have high-order spatial interactions. As a result, this paper calls it a DRSI block. The efficiency of the DRSI block will be demonstrated in the ablation experiments reported in Section 4.6.1.

## 3.4 Cross-Stage Dual-Residual Spatial Interaction Module

Inspired by the bottleneck [45] and CSP structure [48] used in YOLOV5 [33], which has achieved great success in object detection, this study constructed a cross-stage DRSI (C3DR) module based on the DRSI block. The C3DR module consists of standard convolutions and DRSI blocks and has two branches (Figure 5): the main branch and the cross-stage partial branch. The main branch extracts the raw input features solely through a combination of standard convolution, BN and SiLU, while the cross-stage partial branch captures the correlations between features at different stages. The C3DR module integrates the features from the main branch and the cross-stage partial branch in a certain way to achieve feature interaction and integration. Let $x \in \mathbb{R}^{C \times H \times W}$ be the input feature; $x$ will go through two paths: the main branch and the cross-stage partial branch. The starting point of each path is a 1 × 1 standard convolution:

$$x' = \text{Conv}(x), \quad x'' = \text{Conv}(x) \quad x', x'' \in \mathbb{R}^{C_{out}/2 \times H \times W}, \quad (9)$$

where $C_{out}$ represents the final output dimension of the C3DR module; $x'$ and $x''$ are the outputs of the cross-stage partial branch and the main branch, respectively. To reduce the number of parameters and improve the inference speed of the model, after 1 × 1 standard convolutional mapping, the channel dimensions of the cross-stage partial branch and the main branch are reduced to half. As shown in Equation 10, the features of the cross-stage partial branch continue to be input to $n$ dense DRSI blocks:

$$\mathrm{DR}_k(x') = \mathrm{DR}_{k-1}(x'), \quad k = 1, 2, ..., n, \tag{10}$$

where $\mathrm{DR}_k$ is the output of the $k^{\mathrm{th}}$ DRSI block. The cross-stage partial branch fully captures the correlation between different stage features through continuous high-order spatial interactions of $n$ DRSI blocks. Finally, the cross-stage partial branch features extracted by the dense DRSI block are partially integrated with the main branch features in the channel dimension to achieve feature information interactions. Compared to traditional connection methods, the C3DR module can reduce the number of redundant parameters in the network through partial information integration, accelerating the model inference speed. The integrated features are extracted using a standard convolutional layer to obtain the final output result of C3DR, as shown in Equation 11:

$$y = \mathrm{Conv}\left(\mathrm{Concat}\left(x'' + \mathrm{DR}_n(x')\right)\right), \tag{11}$$

where $\mathrm{DR}_n$ is the final output of the cross-stage partial branch after being processed by $n$ consecutive DRSI blocks. The C3DR module has the advantages of a DRSI block and CSP structure. By segmenting the gradient flow and cross-stage feature fusion strategy through the main and cross-stage partial branches, the C3DR module effectively solves the redundant gradient information in the network, reduces the computational load and accelerates the inference speed of the model. Moreover, the continuous high-order spatial interaction of dense DRSI blocks further improves the learning ability of the network. Based on the C3DR module, this paper proposes a novel cross-stage backbone (called CSDR-Net) with high-order spatial interactions, as shown at the bottom of Figure 5.

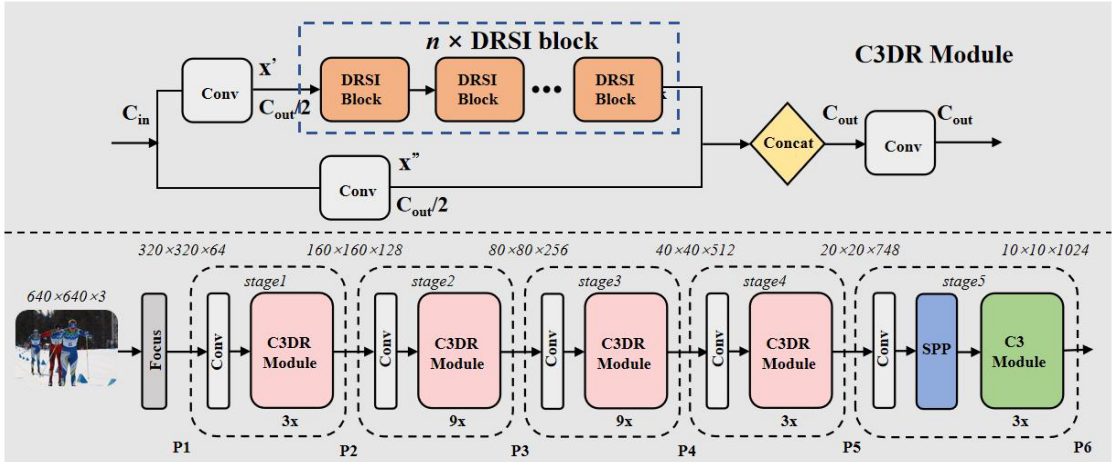

Figure 5: Description of the C3DR module and the CSDR-Net backbone. **Top:** This paper proposes a cross-stage dual-residual spatial interaction module (C3DR) based on the DRSI block. **Bottom:** By stacking C3DR modules, this paper proposes a novel CSDR-Net backbone with high-order spatial interaction capability.

## 3.5 Attentional Spatial Interaction Path Aggregation Network

Herein, two multi-scale feature fusion networks, CBAM-PAN and ASI-PAN, are proposed. The experiments show that they all outperform the original feature fusion network (also known as the path aggregation network(PAN) [49]).

### 3.5.1 CBAM-PAN

The ordinary PAN makes capturing the location rich in context information in the image complex. Hence, this paper introduces the convolutional block attention module (CBAM) [50] into the PAN. Based on PAN, CBAM-PAN replaces all standard convolutions before each upsampling feature in the top–down path with a 1 × 1 CBAM, and a 3 × 3 standard convolution in the bottom–up path is correspondingly replaced with a 3 × 3 CBAM. CBAM-PAN combines the channel attention module (CAM) with the spatial attention module (SAM). CAM can make the network focus on the foreground of the image as well as the meaningful ground-truth regions, while SAM can help the network focus on context-rich positions in the entire image.

The top–down path accepts different scale features extracted from different stages of the backbone network. The feature map with the smallest scale at the top layer is first extracted by standard convolution and reduced in the channel dimension. The extracted features are mapped to a scale consistent with the features of the next layer (the previous stage) through an upsampling operation to achieve multi-scale feature fusion. The fused features are input into the CBAM attention mechanism to capture the context-rich locations in the input image. The fused features will continue the above feature fusion steps to achieve fusion with larger-scale features until the bottom layer. This design enables more detailed information to be passed to higher-level semantic features.

The bottom–up path is the opposite. It relies on downsampling convolution to downsample different scale features processed at different stages of the top–down path, perform feature fusion from large to small scales and hand over to the CBAM attention mechanism for processing. In the bottom–up path, the transfer of high-level semantic information to low-level detail information establishes effective connections between the different levels.

The features of different paths in CBAM-PAN are effectively fused to ensure the full transfer and exchange of information between different levels, thereby avoiding information loss and redundancy. Although CBAM-PAN is very straightforward, experiments have shown that it can considerably improve detection performance.

### 3.5.2 ASI-PAN

Based on CBAM-PAN, this paper further introduces Res-$g^n$Conv into the feature fusion network to construct ASI-PAN. ASI-PAN combines the advantages of Res-$g^n$Conv and CBAM and replaces the 1 × 1 CBAM in the top–down path with Res-$g^n$Conv. Through the continuous high-order spatial interactions of Res-g$^n$Conv in the top–down path and the combination of the dual attention mechanism (CAM and SAM) in the bottom–up path, the network focuses on more meaningful ground-truth regions in the feature map after high-order spatial interactions, as well as context-rich positions. This helps DRSI-Net overcome the feature misalignment issue and detect human bounding boxes and key points more accurately. Experiments show that ASI-PAN outperforms common PAN and CBAM-PAN. They also show that the robust residual spatial interactions of Res-g$^n$Conv can make itself an enhanced alternative to standard convolution.

# 4 Experiments

## 4.1 Dataset and Evaluation Metrics

### 4.1.1 Dataset.

To make better comparisons with state-of-the-art bottom–up methods, the MS COCO dataset [27], which is a challenging and popular benchmark for key point detection, is used to evaluate the proposed method. It is a large dataset with more than 200K images and 250K human instances, including different types of complex samples in MPPE. Each human instance is labelled with the bounding box coordinate and 17 key points. The COCO dataset contains three sets: train2017, val2017 and test-dev2017. The train2017 set comprises 57K images with 150K human instances. The val2017 and test-dev2017 sets consist of 5K and 20K images, respectively. The DRSI-Net was trained on the train2017 set. This paper reports the results on both val2017 and test-dev2017 sets.

### 4.1.2 Evaluation Metrics.

This study follows standard evaluation procedures and uses object key point similarity (OKS) metrics to estimate the COCO pose. The OKS metric is defined as follows:

$$OKS = \frac{\sum_i \exp(-d_i^2/2s^2h_i^2) \cdot \theta(v_i > 0)}{\sum_i \theta(v_i > 0)}, \quad (12)$$

where $d_i$ is the Euclidean distance between the detected key point and the corresponding ground truth; $v_i$ is the visibility flag of the ground truth; $s$ is the object scale; and $h_i$ is the key point constant that controls falloff. $\theta(v_i > 0)$ means if $v_i > 0$ holds, $\theta = 1$, otherwise, $\theta = 0$. In this section, this paper reports average precision and average recall scores with different thresholds and different object sizes: AP (mean of AP scores from OKS = 0.50 to OKS = 0.95 with 0.05 increments), AP50 (AP at OKS = 0.50), AP75, APL for persons of large sizes and AR (mean of AR scores from OKS = 0.50 to OKS = 0.95 with the increment as 0.05).

## 4.2 Implementation Details

The method proposed in this paper adopts a series of data augmentation strategies, including mosaic augmentation with a probability of one, random flip with a probability of 0.5, random translation ([−10, 10]) and random scale ([0.5, 1.5]). Through the anchor selection strategy, three anchors with different proportions are obtained for different scales of persons. These strategies are widely used in object detection. By scaling the width and depth of the entire model to different degrees, three models with almost doubled parameters and complexity are obtained, namely DRSI-Net-s, DRSI-Net-m and DRSI-Net-l. The spatial interaction order of the three models is set to two. The experiment selected different batch sizes for the above three versions of the model. Specifically, the batch sizes of the smallest DRSI-Net-s, medium DRSI-Net-m and largest DRSI-Net-m are set to 36, 16 and 8, respectively. The SGD optimiser with the cosine scheduler is adopted, where the initial learning rate, momentum and decay weight are set to $3 \times 10^{-3}$, 0.937 and $5 \times 10^{-4}$, respectively. The experiment is conducted on Pytorch using four NVIDIA GeForce RTX 3090 GPUs, and the model is trained for 300 epochs. During the test evaluation phase, the large edges of the input image are first adjusted to the desired size to maintain the aspect ratio. To ensure that all input images have the same size, the bottom of the image is filled to generate a square image.

## 4.3 Results on the COCO val2017

Table 1: Comparisons with state-of-the-art methods for the COCO val2017 dataset. All methods used single-scale testing, while † denotes the flipping test.

| Method | Backbone | Input size | #params | GMACS | AP | AP50 | AP75 | APL | AR |
|---|---|---|---|---|---|---|---|---|---|
| EfficientHRNet-$H_0$†[51] | EfficientNetB0 | 512 | 23.3M | 51.2 | 64.8 | 85.2 | 70.7 | 72.8 | 69.6 |
| HigherHRNet†[21] | HRNet-W32 | 512 | 28.6M | 95.8 | 67.1 | 86.2 | 73.0 | 76.1 | |
| HigherHRNet†[21] | HRNet-W32 | 640 | 28.6M | 149.6 | 68.5 | 87.1 | 74.7 | 75.3 | |
| HigherHRNet†[21] | HRNet-W48 | 640 | 63.8M | 308.6 | 69.9 | 87.2 | 76.1 | 76.4 | |
| DEKR†[22] | HRNet-W32 | 512 | 29.6M | 90.8 | 68.0 | 86.7 | 74.5 | 77.7 | 73.0 |
| DEKR†[22] | HRNet-W48 | 640 | 65.7M | 283 | 71.0 | 88.3 | 77.4 | 78.5 | 76.0 |
| YOLOPose-s6[23] | CSPDarknet53-s | 960 | 15.1M | 22.8 | 63.8 | 87.6 | 69.6 | 73.1 | 70.4 |
| YOLOPose-m6[23] | CSPDarknet53-m | 960 | 41.4M | 66.3 | 67.4 | 89.1 | 73.7 | 77.3 | 73.9 |
| YOLOPose-l6[23] | CSPDarknet53-l | 960 | 87.0M | 145.6 | 69.4 | 90.2 | 76.1 | 79.2 | 75.9 |
| DRSI-Net-s | CSDR-Net-s | 960 | 15.4M | 24.6 | 65.3 | 87.8 | 71.5 | 73.5 | 71.7 |
| DRSI-Net-m | CSDR-Net-m | 960 | 36.8M | 64.4 | 68.8 | 89.5 | 74.6 | 78.0 | 74.6 |
| DRSI-Net-l | CSDR-Net-l | 960 | 79.7M | 145.6 | **71.5** | **90.6** | **77.8** | **79.3** | **76.4** |

Table 1 lists the experimental results of the proposed DRSI-Net and other state-of-the-art methods with single-scale testing on the COCO val2017 dataset. DRSI-Net-l achieves the highest

AP of 71.5%, surpassing the results of other state-of-the-art methods. EfficientHRNet [51] is widely used in real-time pose estimations because of its low complexity and good performance. In contrast, the proposed small DRSI-Net-s has higher AP scores and a computational complexity of only 24.6 GMACS. Compared with HigherHRNet [21] and DEKR [22], which are based on the HRNet-W48 backbone, the large DRSI-Net-l has approximately half of the GMACS, but it shows the best performance. Only medium DRSI-Net-m, with 36.8M parameters and almost one-fifth the complexity of DEKR, achieves 89.5% AP50, exceeding the above methods. Compared with the newly proposed YOLOPose [23], DRSI-Net outperforms YOLOPose in all evaluation metrics at the same level of complexity. The performance gain of DRSI-Net on the AP is obvious (Figure 6). On AP50, DRSI-Net-l reached 90.6%. The performance gain of DRSI-Net on the AP50 is relatively small compared to that of YOLOPose, but compared to other state-of-the-art methods, the gain is still obvious.

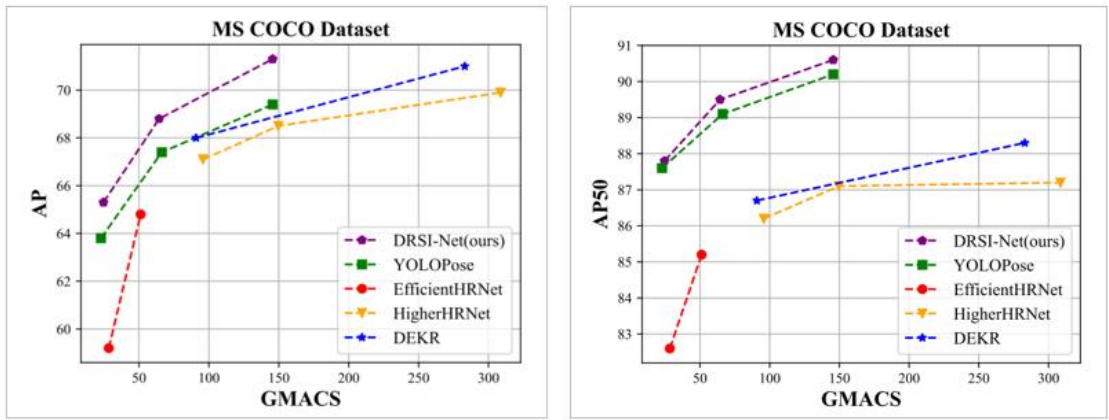

Figure 6: Quantitative comparison with the state-of-the-art methods on the COCO val2017 dataset. **Left:** The proposed DRSI-Net achieves the highest AP scores with half the complexity of the SOTA bottom–up methods. **Right:** DRSI-Net surpasses the SOTA methods on AP50.

## 4.4 Results on the COCO test-dev2017

To better demonstrate the efficiency of the proposed DRSI-Net, more state-of-the-art methods are selected for comparison with the DRSI-Net on the COCO test-dev2017 dataset in Table 2. DRSI-Net with recursive residual spatial interactions outperforms the other methods on the COCO test-dev2017 dataset. With parameters of 15.4M and GMACS of only 24.6, small DRSI-Net-s surpassed the results of OpenPose [16] and Hourglass [52] considerably. With a GMACS of only 145.6, large DRSI-Net-l achieved 70.6% AP, surpassing all the state-of-the-art methods assessed. DRSI-Net has the advantages of YOLOPose and provides outstanding performance on AP50, reaching the highest 90.5% AP50. At the same time, other evaluation metrics of DRSI-Net-l have refreshed the new record, except that AP75 is 0.2 points lower than the higher complexity SOTA DEKR. These results show that the proposed DRSI-Net is a low-complexity and high-accuracy framework for pose estimation that can be a better alternative to existing state-of-the-art bottom–up methods.

Table 2: Comparisons of the COCO test-dev2017 dataset. All methods used single-scale testing, while † denotes the flipping test.

| Method | Backbone | Input size | #params | GMACS | AP | AP50 | AP75 | APL | AR |
|---|---|---|---|---|---|---|---|---|---|
| OpenPose†[16] | | | | | 61.8 | 84.9 | 67.5 | 68.2 | 66.5 |
| Hourglass†[52] | Hourglass | 512 | 277.8M | 413.8 | 56.6 | 81.8 | 61.8 | 67.0 | |
| PersonLab†[53] | ResNet-152 | 1401 | 68.7M | 911 | 66.5 | 88 | 72.6 | 72.3 | 71.0 |
| PiPaf†[17] | | | | | 66.7 | | | 72.9 | |
| HRNet†[12] | HRNet-W32 | 512 | 28.5M | 77.8 | 64.1 | 86.3 | 70.4 | 73.9 | |
| EfficientHRNet-$H_0$†[51] | EfficientNetB0 | 512 | 23.3M | 51.2 | 64.0 | | | | |
| HigherHRNet†[21] | HRNet-W32 | 512 | 28.6M | 95.8 | 66.4 | 87.5 | 72.8 | 74.2 | |
| HigherHRNet†[21] | HRNet-W48 | 640 | 63.8M | 308.6 | 68.4 | 88.2 | 75.1 | 74.2 | |
| DEKR†[22] | HRNet-W32 | 512 | 29.6M | 90.8 | 67.3 | 87.9 | 74.1 | 76.1 | 72.4 |
| DEKR†[22] | HRNet-W48 | 640 | 65.7M | 283 | 70.0 | 89.4 | **77.3** | 76.9 | 75.4 |
| YOLOPose-s6[23] | CSPDarknet53-s | 960 | 15.1M | 22.8 | 62.9 | 87.7 | 69.4 | 71.8 | 69.8 |
| YOLOPose-m6[23] | CSPDarknet53-m | 960 | 41.4M | 66.3 | 66.6 | 89.8 | 73.8 | 75.2 | 73.4 |
| YOLOPose-l6[23] | CSPDarknet53-l | 960 | 87.0M | 145.6 | 68.5 | 90.3 | 74.8 | 76.5 | 75.0 |
| DRSI-Net-s | CSDR-Net-s | 960 | 15.4M | 24.6 | 63.7 | 88.2 | 70.2 | 72.6 | 71.2 |
| DRSI-Net-m | CSDR-Net-m | 960 | 36.8M | 64.4 | 68.1 | 89.8 | 74.4 | 76.5 | 74.0 |
| DRSI-Net-l | CSDR-Net-l | 960 | 79.7M | 145.6 | **70.6** | **90.5** | 77.1 | **76.9** | **75.8** |

## 4.5 Qualitative Comparison with Other Methods

To provide a clearer and more intuitive comparison of the method proposed in this paper with other state-of-the-art methods, DRSI-Net, YOLOPose, DEKR and HRNet were used for pose estimation on the same crowded image containing multiple persons, as shown in Figure 7 (a), (b), (c) and (d). YOLOPose, DEKR and HRNet represent anchor-based, bottom–up and top–down methods, respectively. As shown in Figures 8 (a) and 8 (b), compared to YOLOPose, which is also an anchor-based method, DRSI-Net demonstrates higher accuracy in key point detection. DRSI-Net accurately predicts the occluded left elbow joint and left wrist position of the middle person labelled ① in Figure 7, while YOLOPose exhibited significant deviation in predicting the left wrist position from the ground truth. Figure 7 (c) shows the detection results of the state-of-the-art bottom–up method, DEKR, showing its inferior performance in pose estimation for crowded crowds. Apart from the occluded left elbow joint and left wrist position of the middle person, DEKR also fails to detect numerous key points, as indicated by the red circles in Figure (c). Section 2.3 shows that anchor-based methods can simultaneously detect the bounding boxes and poses of all individuals in a single inference process. Focusing on the bounding box of the middle person detected by DRSI-Net in Figure (a) and the left wrist key point labelled ①, although the left wrist key point labelled ① is outside the defined range of the bounding box, it is predicted correctly by DRSI-Net. Such capability is something that top–down methods cannot achieve. As shown in Figure (d), HRNet missed the detection of the left wrist key point of the middle person. This is because top–down methods require first detecting the bounding boxes of all individuals and then performing key point detection based on these bounding boxes. A comparison of Figures (a) and (d) intuitively demonstrates the difference between anchor-based and top–down methods in this aspect.

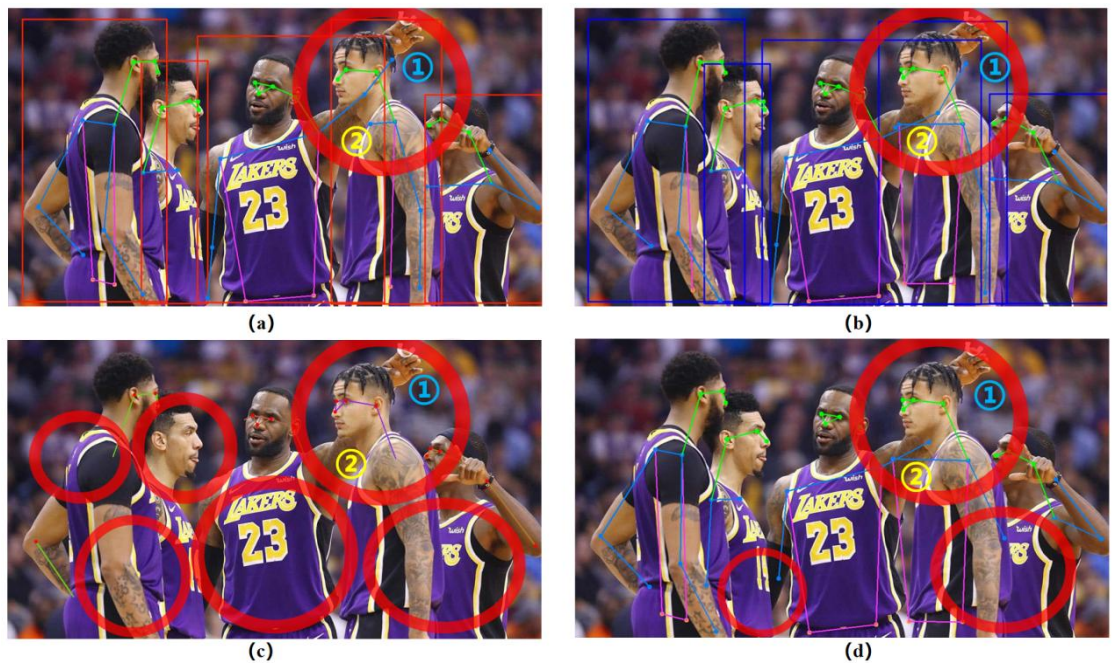

Figure 7: Visual comparison with other state-of-the-art methods. (a) The detection results of DRSI Net-m, YOLOPose-m, DEKR (HRNet-W48) and HRNet-W32 are presented in (b), (c) and (d), respectively.

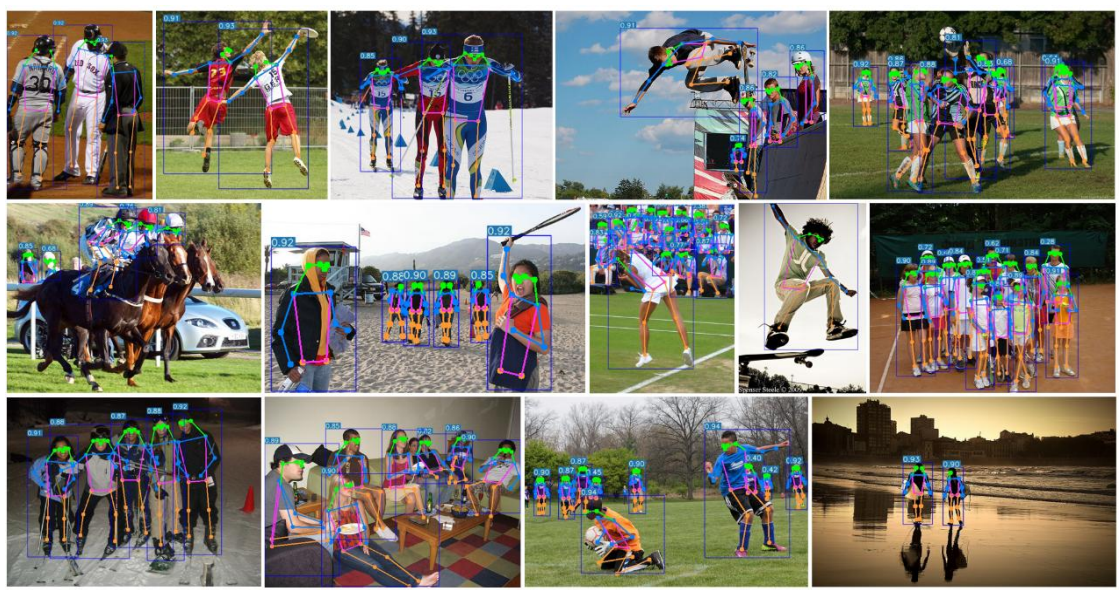

Figure 8: Qualitative results of DRSI-Net. DRSI-Net performs well on a wide range of poses, including occlusions, crowded scenes and viewpoint changes.

Furthermore, as indicated by regions ② labelled (a), (b), (c) and (d) in Figure 7, only the proposed DRSI-Net correctly detects the key points in this region. The other methods either confuse the key points of two individuals or directly fail to detect them. DRSI-Net exhibits better representation capability in complex scenarios such as crowding and occlusion. Figure 8 shows some qualitative results of DRSI-Net. DRSI-Net exhibits robust detection performance in images with different crowding, occlusion, lighting and perspectives.

## 4.6 Ablation Studies

In this section, detailed ablation experiments were conducted to evaluate the proposed DRSI-Net and the effectiveness of each component. Specifically, Section 4.6.1 will experimentally compare the effectiveness of self-attention, $g^nConv$ and Res-$g^nConv$ in the transformer meta-architecture and DRSI meta-architecture, demonstrating the effectiveness of Res-$g^nConv$ and DRSI block. Section 4.6.2 compares the performance of the CSPDarknet53 backbone network (using C3 modules) and the CSDR-Net backbone network (using C3DR modules) with the PAN, CBAM-PAN and ASI-PAN configurations to validate the effectiveness of the C3DR module, CBAM-PAN and ASI-PAN. To visually demonstrate the gain of each component on the proposed DRSI-Net, performance gain results for each component are provided in Section 4.6.3, along with an in-depth analysis. Finally, Section 4.6.4 compares the performance of DRSI-Net and state-of-the-art low-complexity methods [23, 51] at different resolutions. All reported experimental results were evaluated using the COCO val2017 dataset. The experimental results show that the DRSI-Net with recursive residual spatial interactions performs more prominently at low complexity. To reduce the time required for the experiments, the resolution of the input images in the first two ablation experiments was set to 640.

### 4.6.1 Res-$g^nConv$ vs. GnConv and DRSI Block Meta-architecture vs. Transformer Block Meta-architecture.

To prove the effectiveness of Res-$g^nConv$ (proposed in Section 3.1) and the DRSI block (proposed in Section 3.2), this paper designed the comparative experiment shown in Table 3. The result of incorporating the transformer block [26] into DRSI-Net-s is reported for the first time. Method 1, based on the transformer block, showed a 'CUDA out-of-memory' phenomenon due to the high quadratic complexity of self-attention. The Hor block [24] has the same meta-architecture as the transformer block, only replacing self-attention with $g^nConv$ with similar spatial interactions. Compared with Method 1, Method 2, based on the Hor block, does not appear to burst the memory. This shows that $g^nConv$, based on standard convolution, effectively avoids the high quadratic complexity of self-attention. The R-Hor block is designed, and Method 3 is developed based on it to prove the effectiveness of the proposed Res-$g^nConv$. The R-Hor block has the same meta-architecture as the transformer block and Hor block, only replacing the position of self-attention and $g^nConv$ with a Res-$g^nConv$. Compared to Method 2 based on the Hor block (based on $g^nConv$), Method 3 composed of the R-Hor block (based on Res-$g^nConv$) shows no increase in complexity; however, all evaluation metrics have been improved. The experimental results effectively demonstrate that Res-$g^nConv$ is a more robust and effective convolution with residual spatial interactions.

Table 3: Res-gnConv vs. gnConv and DRSI block meta-architecture vs. transformer block meta-architecture.

| Method | Block | Transformer meta-architecture | DRSI meta-architecture | self-attention | $g^nConv$ | Res-$g^nConv$ | GMACS | AP | AP50 | AP75 |
|---|---|---|---|---|---|---|---|---|---|---|
| 1 | Transformer | √ | | √ | | | CUDA out of memory | | | |
| 2 | Hor | √ | | | √ | | 10.8 | 57.6 | 84.0 | 61.2 |
| 3 | R-Hor | √ | | | | √ | 10.8 | 58.3 | 84.4 | 62.5 |
| 4 | **DRSI** | | √ | | | √ | **10.9** | **59.5** | **85.0** | **64.6** |

Unlike the transformer block meta-architecture adopted in the above three blocks, the proposed DRSI block uses a novel meta-architecture. A new inverted bottleneck and Res-$g^nConv$ together constitute the DRSI block. In the last row of Table 3, this paper reports the results of DRSI block-based Method 4 (i.e. proposed DRSI-Net-s). Method 4 achieved 59.5% AP, 85% AP50 and 64.6% AP75 on the COCO val2017 dataset, which surpasses the results of the R-Hor block-based Method 3 (based on the transformer block meta-architecture and Res-$g^nConv$). This result demonstrates the effectiveness of the DRSI block. Through the above ablation experiment, Res-$g^nConv$ and DRSI block meta-architecture make noticeable contributions to the overall model performance, and better results can be obtained using these two structures together.

## 4.6.2 Comparisons of Different Backbones and Different Path Aggregation Networks

Table 4: Comparisons of the different backbones and different path aggregation networks.

| Method | CSPDarknet53 | CSDR-Net | PAN | CBAM-PAN | ASI-PAN | GMACS | AP | AP50 | AP75 |
|---|---|---|---|---|---|---|---|---|---|
| 1 | √ | | √ | | | 10.3 | 57.5 | 84.3 | 61.1 |
| 2 | √ | | | √ | | 10.2 | 58.6 | 84.8 | 62.6 |
| 3 | √ | | | | √ | 11.1 | 58.6 | 84.7 | 62.9 |
| 4 | | √ | √ | | | 10.9 | 58.7 | **85.4** | 62.7 |
| 5 | | √ | | √ | | 10.9 | 59.5 | 85.0 | 64.6 |
| 6 | | √ | | | √ | 11.8 | **59.6** | 84.8 | **64.8** |

In Sections 3.4 and 3.5, the C3DR module, CSDR-Net backbone and two new PANs, CBAM-PAN and ASI-PAN, are proposed. This section verifies the effectiveness of the above components through the comparative experiments in Table 3. Method 1 (i.e. YOLOPose) comprises a CSPDarknet53 [48] backbone (based on the C3 module) and an ordinary PAN [49]. A comparison of the results of Methods 1 and 4 on the COCO val2017 dataset indicates that Method 4, composed of the CSDR-Net backbone (based on the C3DR module) and PAN, has 1.2% AP, 1.1% AP50 and 1.6% AP75 improvement, and their complexity is similar. Method 4 achieved the highest AP50 of 85.4% only by replacing the backbone with CSDR-Net, surpassing the other methods. Therefore, the proposed C3DR module is efficient and lightweight. Using CSDR-Net with effective residual spatial interactions as a backbone can also improve the representation ability of the network.

This study examines the integration of channel and spatial attention into the features after high-order residual spatial interactions, helping the network focus on more meaningful ground-truth regions and positions rich in context information to fuse multi-scale features more efficiently and refine the generated poses. A comparison of Methods 1, 2 and 3 indicates that the proposed CBAM-PAN and ASI-PAN outperformed PAN. A comparison of Methods 1 and 5 shows that AP, AP50 and AP75 are increased by 2.0%, 0.7% and as much as 3.5%, respectively, when the CSDR-Net backbone and CBAM-PAN are combined. Combining the CSDR-Net backbone and ASI-PAN in Method 6 further improves the AP and AP75 scores.

### 4.6.3 Contribution of Different Components to the Network

Table 5: Comparison of the contribution of different components to the model on the COCO val2017 dataset.

| Method | GMACS | AP | AP50 | AP75 |
|---|---|---|---|---|
| Base | 10.3 | 57.5 | 84.3 | 61.1 |
| Base + **Res-$g^n$Conv** | 10.5 | 58.0 | 84.6 | 62.0 |
| Base + Res-$g^n$Conv + **DRSI block** | 12.7 | 58.6 | 84.8 | 62.3 |
| Base + Res-$g^n$Conv + DRSI block + **C3DR** | 10.9 | 58.7 | **85.4** | 62.7 |
| Base + Res-$g^n$Conv + DRSI block + C3DR + **CBAM-PAN** | 10.9 | 59.5 | 85.0 | 64.6 |
| Base + Res-$g^n$Conv + DRSI block + C3DR + **ASI-PAN** | 11.8 | **59.6** | 84.8 | **64.8** |

This study used YOLOPose as the baseline to compare the contributions of different components proposed in this paper to the model. The experimental results are obtained after adding the different components. As shown in Table 5, after introducing Res-$g^n$Conv, the AP, AP50 and AP75 increased by 0.5%, 0.3% and 0.9%, respectively, with only an increase of 0.2 GMACS. Hence, Res-$g^n$Conv has a positive effect on the performance of the model. After introducing the DRSI block, although the complexity of the model increased significantly by 2.2 GMACS, the performance of the model also improved considerably. The combination of Res-$g^n$Conv and DRSI block improved the model performance by 1.1%AP, 0.5%AP50 and 1.2%AP75 compared with the baseline. After introducing the C3DR module, the complexity of the model decreased significantly by 1.8 GMACS compared to the model using only the DRSI block, and AP, AP50 and AP75 were improved, with AP50 showing the most significant improvement, reaching a maximum of 85.4%. Therefore, the C3DR module can optimise the computational efficiency and performance of the model. The last two rows of Table 5 show the gains achieved by replacing the ordinary PAN with two new path aggregation networks proposed in this paper: CBAM-PAN and ASI-PAN. The experimental models, except for the last two rows, use PAN. The addition of CBAM-PAN significantly improved AP and AP75 of the model. Compared to the baseline, using CBAM-PAN increased AP and AP75 by 2.0% and 3.5%, respectively. Thus, CBAM-PAN has a significant benefit in improving the performance of the model. The introduction of ASI-PAN further improved the AP and AP75 of the model. These experimental results and analysis fully demonstrate the effectiveness of several innovative components proposed in this paper.

### 4.6.4 Across Resolution

Table 6: Comparison with SOTA low complexity (less than 50 GMACS) methods on the COCO val2017 dataset.

| Method | Input | GMACS | AP | AP50 |
|---|---|---|---|---|
| DRSI-Net-s | 1280 | 43.7 | **66.0** | **88.7** |
| DRSI-Net-s | 960 | 24.6 | **65.3** | 87.8 |
| DRSI-Net-s | 640 | 10.9 | 59.5 | 85.0 |
| DRSI-Net-s | 512 | 7.0 | 54.9 | 82.1 |
| DRSI-Net-s | 448 | 5.3 | 51.4 | 80.2 |
| DRSI-Net-s | 384 | 3.9 | 47.3 | 75.9 |
| YOLOPose-s | 1280 | 40.6 | 64.9 | 88.4 |
| YOLOPose-s | 960 | 22.8 | 63.8 | 87.6 |
| YOLOPose-s | 640 | 10.1 | 57.5 | 84.3 |
| YOLOPose-s | 512 | 6.5 | 52.3 | 80.9 |
| YOLOPose-s | 448 | 4.9 | 49.0 | 78.0 |
| YOLOPose-s | 384 | 3.6 | 44.9 | 74.2 |
| EfficientHRNet-H$_{-1}$ | 480 | 28.4 | 59.2 | 82.6 |
| EfficientHRNet-H$_{-2}$ | 448 | 15.4 | 52.9 | 80.5 |
| EfficientHRNet-H$_{-3}$ | 416 | 8.4 | 44.8 | 76.7 |
| EfficientHRNet-H$_{-4}$ | 384 | 4.2 | 35.7 | 69.6 |

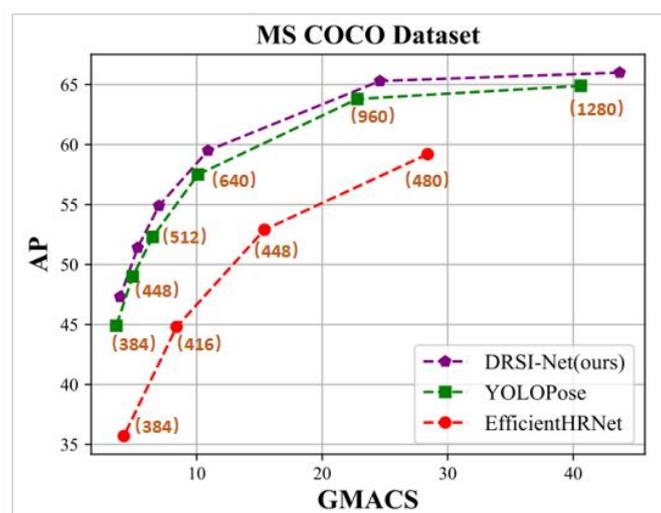

Figure 9: AP vs. GMACS for low-complexity methods (less than 50 GMACS) on the COCO val2017 dataset.

The proposed method is performed across various resolutions. As shown in Table 6, when the resolution is 960, DRSI-Net-s reached 65.3% AP, which cannot be achieved with YOLOPose-s [23] even at 1280. The performance gain of the AP is greatly reduced when the resolution is increased to 1280. The base resolution is set to 960 because DRSI-Net has achieved competitive performance at a resolution of 960.

As shown in Figure 9, the AP scores of DRSI-Net-s on the COCO val2017 dataset are higher than those of existing state-of-the-art low-complexity methods, such as YOLOPose and EfficientHRNet. The good performance obtained in the low-complexity cases shows that the proposed DRSI-Net is a framework with low complexity and high accuracy, making it suitable for real-time MPPE.

# 5 Conclusion

In this paper, a novel framework called DRSI-Net for MPPE is proposed. By performing continuous residual spatial interactions, the framework captures long-range dependencies between features, which effectively improve the detection accuracy of human key points. This paper proposes three components with high-order spatial interactions, i.e. Res-g$^n$Conv, DRSI block and the cross-stage DRSI module. These components are lightweight and plug-and-play. They can be easily inserted into the network to improve the spatial interaction and representation abilities of the entire network. Furthermore, integrating spatial interactions and channel/spatial dual attention mechanisms in the path aggregation network helps the network adaptively focus on the features relevant to the target key points. Thus, it effectively fuses multi-scale features and further refines the generated poses. The effectiveness of the above components has been verified in ablation experiments. Together, they provide the proposed DRSI-Net low-complexity and high-accuracy framework for MPPE. The experimental results on the challenging COCO dataset also provide evidence of this.

Although DRSI-Net is designed for MPPEs, its core spatial interactions have many potential applications, such as object detection and semantic segmentation. A future study will explore the potential of residual spatial interactions in more general applications. In addition, future studies will combine human–human interactions and high-order spatial interactions to improve the performance of MPPEs.